\documentclass{article}

\usepackage[preprint,nonatbib]{nips_2018}

\usepackage[utf8]{inputenc} 
\usepackage[T1]{fontenc}    
\usepackage{url}            
\usepackage{booktabs}       
\usepackage{amsfonts}       
\usepackage{nicefrac}       
\usepackage{microtype}      
\usepackage{comment}
\usepackage{graphicx}
\usepackage{subfig}
\usepackage{tabu}
\usepackage{color}

\usepackage{floatrow}
\graphicspath{{Figures/}}

\title{Flow Shape Design for Microfluidic Devices Using Deep Reinforcement Learning}

\author{
  Xian Yeow Lee\\
  Iowa State University\\
  \texttt{xylee@iastate.edu} \\
  \And
  Aditya Balu \\
  Iowa State University\\
  \texttt{baditya@iastate.edu} \\
  \AND
  Daniel Stoecklein \\
  University of California, Los Angeles\\
  \texttt{stoeckd@g.ucla.edu} \\
  \And
  Baskar Ganapathysubramanian \\
  Iowa State University\\
  \texttt{baskarg@iastate.edu} \\
  \And
  Soumik Sarkar \\
  Iowa State University\\
  \texttt{soumiks@iastate.edu} \\
}

\begin{document}

\maketitle

\begin{abstract}
Microfluidic devices are utilized to control and direct flow behavior in a wide variety of applications, particularly in medical diagnostics. A particularly popular form of microfluidics -- called inertial microfluidic flow sculpting -- involves placing a sequence of pillars to controllably deform an initial flow field into a desired one. Inertial flow sculpting can be formally defined as an inverse problem, where one identifies a sequence of pillars (chosen, with replacement, from a finite set of pillars, each of which produce a specific transformation) whose composite transformation results in a user-defined desired transformation. Endemic to most such problems in engineering, inverse problems are usually quite computationally intractable, with most traditional approaches based on search and optimization strategies. In this paper, we pose this inverse problem as a Reinforcement Learning (RL) problem. We train a DoubleDQN agent to learn from this environment. The results suggest that learning is possible using a DoubleDQN model with the success frequency reaching 90\% in 200,000 episodes and the rewards converging. While most of the results are obtained by fixing a particular target flow shape to simplify the learning problem, we later demonstrate how to transfer the learning of an agent based on one target shape to another, i.e. from one design to another and thus be useful for a generic design of a flow shape.
\end{abstract}

\section{Introduction}

The field of microfluidics has been a popular locus of study for biological, chemical, and manufacturing research communities for nearly two decades, largely due to  modern fabrication methods making the small length scales ($\sim O(1) - O(100)$ $\mu$m) more accessible to high-precision analysis for biomedical technology, microscale engineering, and the study of fundamental physical systems~\cite{Whitesides2006, Sackmann2014}. Microfluidic flows are typically dominated by viscous forces, as characterized by the Reynolds number, $Re=\frac{\rho V D_H}{\mu}$ (the ratio of inertial forces to viscous forces in a fluid, with fluid density $\rho$, velocity $V$, viscosity $\mu$, and microchannel hydraulic diameter $D_H$), which becomes $Re\approx0$ due to small $D_H$ in microfluidic devices. In the last decade, however, inertially dominated flows ($Re > 1$, often via increased fluid velocity $V$) have seen a rapid increase in attention due to their interesting nonlinear behavior~\cite{Amini2014}. One such effect is seen in fluid flowing past an obstacle within a microchannel, as the finite inertia breaks the fore-aft symmetry around the obstacle, irreversibly deforming the fluid. Amini et al.~\cite{amini2013engineering} leveraged this effect by sequencing cylindrical pillars in a programmatic way to ``sculpt'' the flow. Stoecklein et al.~\cite{labonachip} used an efficient forward model for simulating inertial flow sculpting to explore different flow shape deformations, illustrating a diverse and richly populated design space with many different possible flow shapes. 

The ability to engineer the cross-sectional shape of fluid streams has since been used to fabricate tailored microfibers~\cite{Nunes2014} and 3D microparticles~\cite{Paulsen2015, Wu2015, Paulsen2018}, perform solution exchange~\cite{Sollier2015}, and create 3D shaped microcarriers for cellular analysis~\cite{Wu2018}. Flow sculpting shows additional promise for creating useful micro-structures for tissue engineering~\cite{Du2008}, biosensing~\cite{LeGoff2015}, as well as diagnostics and drug delivery~\cite{Yu2018}. But design in flow sculpting is extremely difficult: even the limited set of 32 pillar configurations used by Stoecklein et al.~\cite{labonachip} offered $>10^{15}$ pillar sequence permutations, with recent developments providing a continuous set of infinite pillar geometries~\cite{Stoecklein2018}, vastly increasing flow shape possibilities, but making manual design impossible. This motivates the need for a fast solution to the inverse problem in flow sculpting, which asks: {\it given a desired fluid flow shape, what is the pillar sequence which will create the closest matching shape?} 

The idea of designing shapes from a microfluidic flow is not new. In \cite{gaoptimization}, the inverse problem is posed as an optimization problems of choosing from a sequence of 32 pillar configurations. The authors utilize the Markov property of the pillar transformations -- i.e. only the outcome of the previous pillar affects the subsequent pillar -- to formulate an optimization approach based on Genetic Algorithms. While successful, this approach is quite time consuming taking several dozens of minutes for a solution, thus precluding real-time design exploration which is especially crucial in the bioengineering community for device design. For such problems with an combinatorially large exploration space, it is natural to use Reinforcement Learning (RL). The RL community has made significant advances over the past few years, and we leverage successes such as DQN~\cite{DQN}, Double DQN~\cite{DDQN} to solve the design problem. There exist many works on RL which are interesting in this context but in this study we restrict ourselves to Double DQN~\cite{DDQN}. In future works, we would explore several other algorithms to improve the performance of the inverse design problem. 

\subsection{Related Work}
The initial attempts for this design problem was done by Stoecklein et.al.\cite{gaoptimization}  and Amini et. al.\cite{amini2013engineering} using Genetic Algorithm (GA) for solving this inverse design problem. The key disadvantage with such an approach is that the search space in a GA based approach depends on the size of the pillar sequence chosen and search space for a pillar sequence of size $s$ is $32^s$ sequences. Such a large search space is very difficult to explore, which naturally makes GA a very time consuming option. Further, this approach requires the same exploration to be performed any time a new design problem is being solved. Also, multiple trials for each configuration required to ensure statistical convergence of optima compounds the cost even further. 

In this context, several deep learning based approaches were attempted. Lore et. al~\cite{smccnn} show that it is possible to combine a convolutional neural network (CNN) with a simultaneous multi-class classification (SMC) problem formulation to predict a fixed length sequence of pillars given a target flow shape. In this formulation, the network predicts the pillar type for each index of the sequence simultaneously using a modified loss function which sums the negative log-likelihood across each pillar in the predicted sequence. However, the performance of this model depended heavily on the sample set used for generation. In this context, Stoecklein et.al~\cite{scirep} explored the sampling complexity of the data generation by performing PCA and HDMR and also understood the need for a non-uniform sampling to solve this problem. Further, Lore et. al~\cite{ppnitn} explored another approach in representing the problem using two networks, (i) Pillar prediction network and (ii) Intermediate transformation network. The idea is to subdivide the inverse problem by first identifying the intermediate shapes using the intermediate transformation network and then use pillar prediction network to predict the pillar sequence. However, the only knob controlling when to subdivide is based on the parameter called complexity measure. In reality, this methodology ends up generating pillar sequences that are longer compared to the actual pillar sequence. This indicates the inefficiency in representing the problem appropriately. Additionally, having a longer pillar sequence also implies higher cost for construction of the microfluidic device. Hence, obtaining pillar sequences which are optimal in length and naturally adaptive to sampling and performing better than evolutionary algorithms is necessary. In particular, learning such a design domain is useful for design problems where we need results in real time. Therefore, it is infeasible to explore the search space of the domain repeatedly for every design problem as performed by an evolutionary algorithm.

Our approach is also motivated with several other related works. In \cite{manochasiggraph18}, RL was leveraged to control the flow direction in an application for rigid body control. In \cite{stephanguy}, RL was used to produce safer and shorter trajectories than traditional motion planning. In \cite{rlvortexflow, rlvortexflow2}, they attempted to use RL for control over vortex induced flows. In \cite{rlgliding}, RL was applied to achieve gliding with either minimum energy expenditure, or fastest time of arrival, at a predetermined location. However, to the best of the knowledge of the authors, there are none or very sparse work in the area of Flow-Shape design for microfluidic devices. 

\subsection{Contributions}
In this work we build a generic framework for solving engineering design problems using RL. Our specific contributions are:
\begin{itemize}
    \item We build an RL environment using OpenAI gym~\cite{gym} for addressing the flow shape design in a microfluidic device.
    \item We further build Double DQN agents for solving the design problem for obtaining few key shapes as a proof-of-concept.  
\end{itemize}
The flow of the rest of the paper is as follows: In section~\ref{sec:flowsculpt}, we explain the details about the design environment. In section~\ref{sec:model}, we use the design environment constructed with Double DQN algorithm and then learn the RL environment. Further, in section~\ref{sec:results}, we show some preliminary results obtained and also show how transfer learning is helpful in this context. Finally, we share some conclusions and future work in section~\ref{sec:conclusions}.

\section{Flow Shape Design}\label{sec:flowsculpt}
We formally pose the flow-sculpting problem (i.e., obtaining the pillar sequence to reach a target flow shape) as a deep reinforcement learning problem. In the RL problem, an agent interacts with the environment that contains certain dynamics governed by the physics of the real world. For each time step $t$, the agent receives a state $s_t$ from the environment and selects an action $a_t$ according to it's policy function $\pi(s|a)$. Here, we use a deep neural network for the policy that represents the mapping from the state space to the action space. The agent then receives a scalar reward $r_t$ computed from a reward function $R(s)$ and transitions to the next state $s_{t+1}$ based on dynamics of the environment. The process continues until the maximum episode length is reach or the goal of the RL agent is met.

\begin{figure}[!h]
    \centering
    \includegraphics[width=1\linewidth, clip, trim={0 1.5in 0 1.5in}]{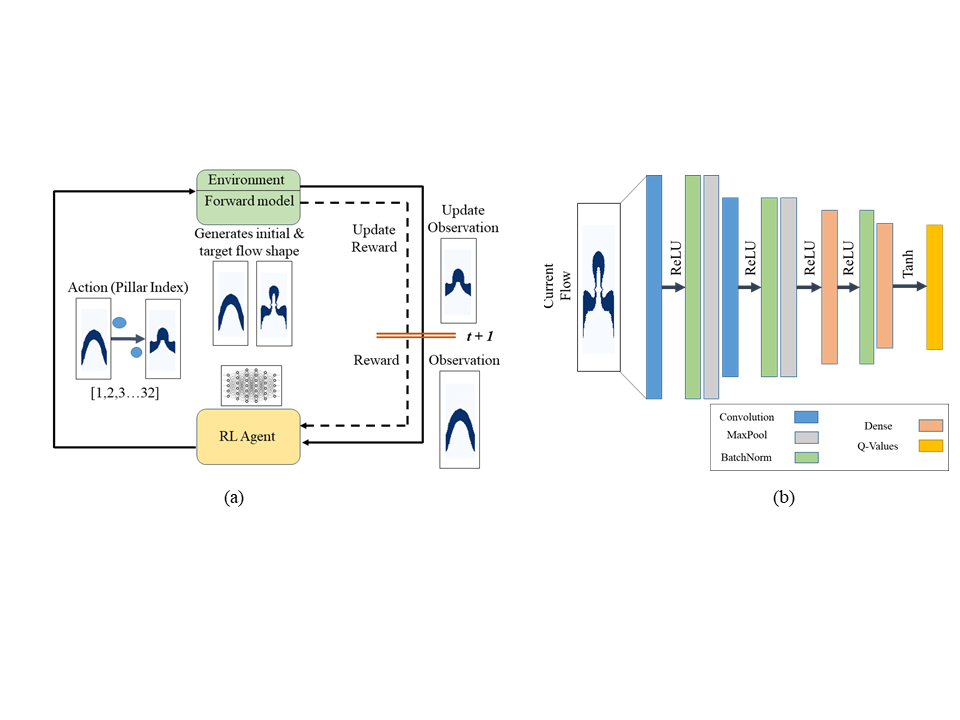}
    \caption{Deep Reinforcement Learning Framework for flow shape design: (a) Formulation of how to perform a reinforcement learning for design of flow shape, the current flow shape is the state/observations. The action is the taken by the RL Agent and environment computes the next flow shape and a reward is computed for the action. (b) The agent policy takes the current flow shape and provides the prediction of action with maximal reward.}
    \label{fig:Network}
\end{figure}

We embed the Markovian process of moving from one flow shape to another using the transition matrix as a RL problem using an OpenAI gym environment (as shown in Fig.~\ref{fig:Network}). Since the dynamics of the environment is represented by a set of equations, the environment is deterministic. We consider that the current flow shape as the fully observable state of the environment and therefore we make no distinction between the observation and state returned by the environment. For each new episode, the target flow shape is initialized and given as the goal for the agent.
At each time step $t$, the RL agent is tasked with choosing the best action corresponding to the different possible pillar configurations. The action is passed to the environment where the corresponding transformed flow shape for the current flow shape is computed using the markovian transition matrix discussed in \cite{gaoptimization}. This transformed shape is considered as the new observation for the agent. The episode terminates only when the agent places a sequence of pillars which deforms the flow shape to a shape that is similar to target flow shape or when the number of pillars placed exceeds a certain constraint.  A metric (Pixel Match Rate, PMR) is defined as the $l1$-norm of each pixel between the flow shape generated by the agent and the target flow shape. The PMR ranges from 0 where no pixels match between the target and generated flow shape to 1 if both the target and generated flow shape are identical for each pixel. Thus, the goal is achieved if a threshold of the PMR is reached.

\begin{equation}
    PMR = 1 - \frac{|I_t-I_p|}{I_t}
    \label{Eq:PMR}
\end{equation}

Since the structure of the reward is instrumental to the learning capabilities of the RL agent, we shaped the reward as function of the PMR. To incentivise the agent to achieve the target flow shape with the least amount of pillars, we structure the reward function in the gym environment to be negative reward. That is, for each step the agent takes, it receives a reward equivalent to the PMR difference between the flow shape it generated and the target flow shape, scaled by some constant. We find that the reward function shown in Equation~\ref{Eq:Reward}, which is shaped by subtracting the baseline $b$ of the worse case scenario and scaled by the remainder of the PMR range after subtracting the baseline, is rich enough for the agent to learn the dynamics of environment and achieve the target flow shape. Here, we empirically find that the general baseline for the PMR metric is approximately 0.5 from running mutliple experiments with different target flow shapes.

\begin{equation}
    Reward = -(1 - \frac{PMR-b}{b})
    \label{Eq:Reward}
\end{equation}

In formulating this problem as a reinforcement learning problem, we consider two scenarios. In the first scenario, we generate a target flow shape using a sequence of pillars defined by the user. In this scenario, the target flow shape remains the same throughout the agent's training process and is analogous to the agent learning to navigate to a single goal. We ensure that the agent explores the action space and finds multiple possible paths (i.e. designs of pillar sequences) to reach the target (i.e. the cross-section shape of the flow).
\\
\\
We then formulate a second problem as a transfer reinforcement learning problem. In this formulation, we first train a RL agent on a flow shape that is easily achievable. Using the parameters of the learned agent, we then train the agent on sequence of flow shapes that are increasingly harder to achieve. We hope that by transferring the learned parameters of the agent, the more complex flow shapes can be more easily learned by the agent in a shorter time or with less exploration or both. 

\section{Algorithms and Model} \label{sec:model}
To demonstrate that RL agent can learn a broad spectrum of flow shapes, we randomly sample flows from a set of 148,026 pillar sequences that was previously generated in~\cite{scirep}. These pillar sequences were generated from 7.5 million random sequences with 32 possible pillar configurations. A PCA was performed on the resulting 7.5 million flow shapes to produce a dataset of 148,026 unique flow shapes. Thus, we demonstrate that the agent in a fixed-flow environment can sufficiently learn a target flow shape for a wide range of flow shapes. Since, the 10 sequences we sampled have a pillar sequence length of 7, we set the maximum episodic length of our environment to be 7 so that the RL agent either reach an approximation of the target flow shape using pillar sequence of length 7 or less. The action space of the agent is a set of 32 discrete actions corresponding to the 32 possible pillar configurations. 

To test the efficacy of our proposed framework, we used a DDQN algorithm on the flow-sculpting environment because value iteration methods are known to be more sample efficient compared to the policy iteration methods~\cite{DDQN}. In DQN~\cite{DQN}, the deep neural network acting as the Q-function approximator approximates the Q-value for each state-action pair. The RL agent then selects the action with the maximum Q-value. The DDQN algorithm is an improvement over the DQN introduced by \cite{DDQN}. In essence, the DDQN algorithm reduces the over estimation of a state's value by decoupling the network used to select the next best action and the network used to estimate the value of the state. 
The policy of the DDQN agent is represented by a deep convolutional neural network with 2 convolutional layers with 32 and 64 output filters respectively. Each convolution layer is also followed by a subsequent layer of batch normalization and max-pooling. After the convolution layers, we use two fully-connected layers with 128 and 64 hidden units respectively and a batch normalization layer in between them to give an output of the Q-values for the 32 possible actions.

The termination criteria that we use for the agent was Pixel-Match-Rate (PMR). A 90\% PMR is set as the termination threshold if the finer details of the target flow shape is required. However, if the application of the microfluidic requires just a bulk approximation of the target flow shape, 85\% PMR is typically sufficient~\cite{scirep}.

We also used RMSProp with a learning rate of 0.001 as the optimizer for the agent and a linearly decaying exploration policy which decays from 1 to 0.1 over 1 million steps. The target network update interval was set at $4000$ steps and the update only begins after the agent has taken $10K$ random steps. For each our sampled flow shape, the agent is trained for $300K$ episodes. 

\section{Results and Discussion} \label{sec:results}
\begin{figure}[h!]
 \centering
 \includegraphics[width=1\linewidth,clip,trim={0in 1.5in 0in 2in}]{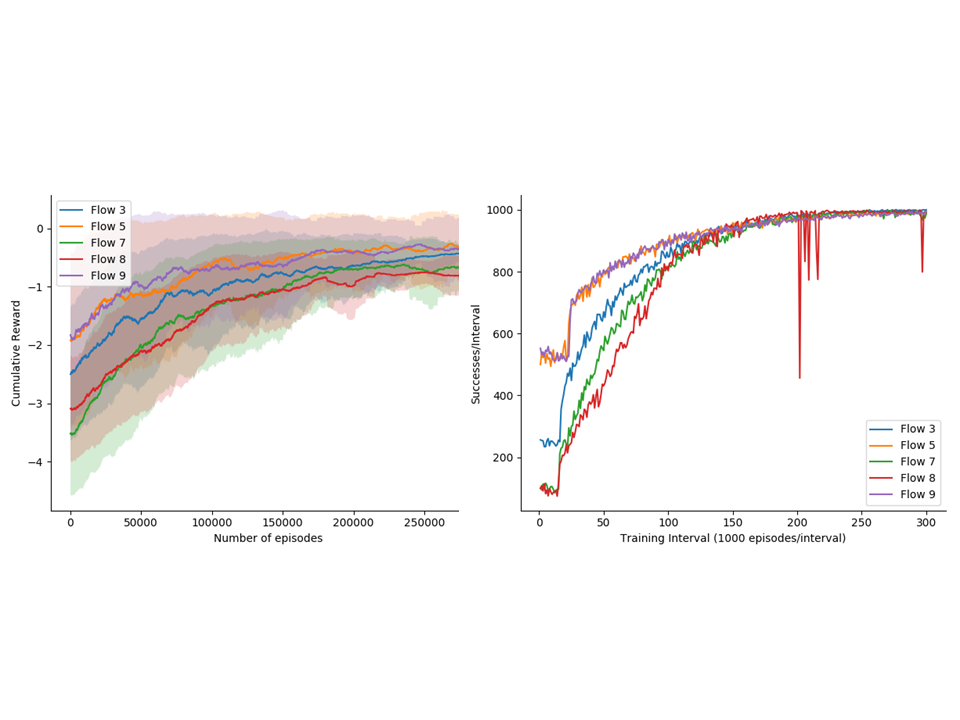}
 \caption{Reward and Frequency of Success: We show here the performance of the agents on 5 flows. The cumulative reward for each episode is shown with respect to the number of episodes (left). The frequency of reaching the success of being able to reach the terminal state of success is shown for an interval of 1000 episodes.}
 \label{Fig:Rewards}
\end{figure}

In Figure~\ref{Fig:Rewards}(a), we show the average cumulative reward achieved by the agent for 5 of the target flow shapes (Flow ID 3, 5, 7, 8 and 9) where the agent performed well using a threshold of 90\% PMR. As seen in the figure, the cumulative rewards attained by the RL agent converges after $300K$ episodes of training. This convergence is similarly reflected in Figure 2(b) where the RL agent's frequency of success per 1000 episode is shown. It can be seen that the RL agent achieves almost a success rate of 100 percent for all of the flows. 

\begin{figure}[!t]
 \centering
 \includegraphics[width=1\linewidth,clip,trim={0in 0in 0in 0in}]{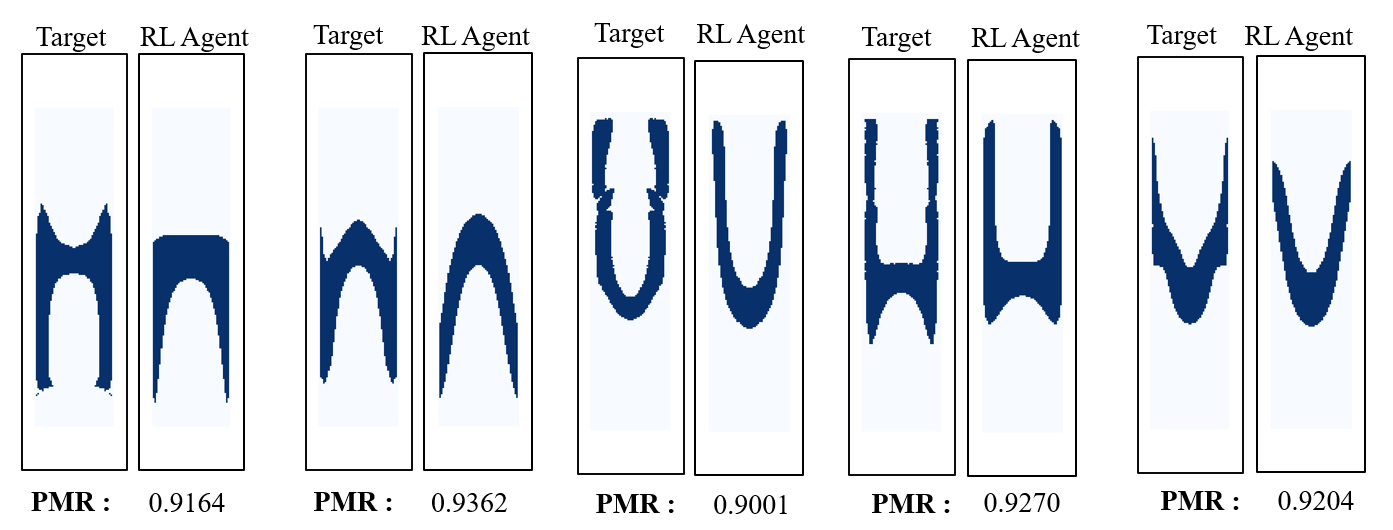}
 \caption{Comparison of target flow shapes, flow shapes generated by RL agent and their corresponding pixel match rates (PMR)}
 \label{Fig:Comparisons}
\end{figure}

\begin{table}[h!]
\caption{True sequence of target flow shapes and sequence predicted by RL agent}
\begin{tabular}{l l l}
 \hline
  Flow Shape ID & Target Sequence & RL Agent \\
 \hline
 Flow 3  & 15, 25, 6, 20, 18, 13, 15  & 24, 23  \\ 
 Flow 5  & 22, 11, 31, 15, 18, 4, 1  & 25  \\ 
 Flow 7  & 22, 29, 24, 16, 13, 12 , 4  & 29, 30  \\
 Flow 8  & 20, 14, 16, 23, 17, 4 , 6  & 30,30,24  \\ 
 Flow 9  & 23, 6, 3, 5, 5, 14, 11  & 20  \\
\hline
\label{Tab:TrueVsAgent}
\end{tabular}
\end{table}

A comparison of the target flow shapes and flow shapes generated by the RL agent is shown in Figure~\ref{Fig:Comparisons}. An interesting observation here is that the flow shapes sculpted by the RL agent were all generated with a minimal number of pillars. While the target flow shapes were all generated using a sequence of 7 pillars, the longest sequence of pillars used by the RL agent was 3 with a majority of the shapes being achievable with 2 pillars or less. Besides the obvious benefit of being more cost efficient, this can be also extremely advantageous in a microfluidic setting as often times there are physical constraints which limits the size of a microfluidic device. Therefore, the ability to generate a bulk approximation of the target flow shape with significantly shorter pillar sequence is extremely exciting. Although ultimately, it is up to the user of the microfluidic device to specify the threshold of accuracy required for his/her respective application.

We also demonstrate that it is possible for the RL agent to arrive at multiple possible solutions for the same inverse problem. We use the example of Flow 5 to illustrate this occurrence. Table~\ref{Tab:Best 5} shows the top 5 pillar sequences generated by the RL agent where the resulting deformed flow shapes has at least 90 percent PMR with the target flow. This can also be leveraged greatly in multiple ways. First, the user of the microfluidic device can utilize the RL agent as a recommender system. If the practitioner of the microfluidic device has a general idea of the required flow shape, the flow shape can be first used as target flow shape and the RL agent can be used to further fine-tuned the structure of the flow through the multiple solutions proposed by the agent. Secondly, if the user only has certain types of pillars at his/her disposal due to various factors, such as manufacturing capability constraints, this can be used to generate an alternate solution which might be more feasible for the user.

\begin{figure}[h!]
 \centering
 \includegraphics[width=0.6\linewidth,clip,trim={0in 0in 0in 0in}]{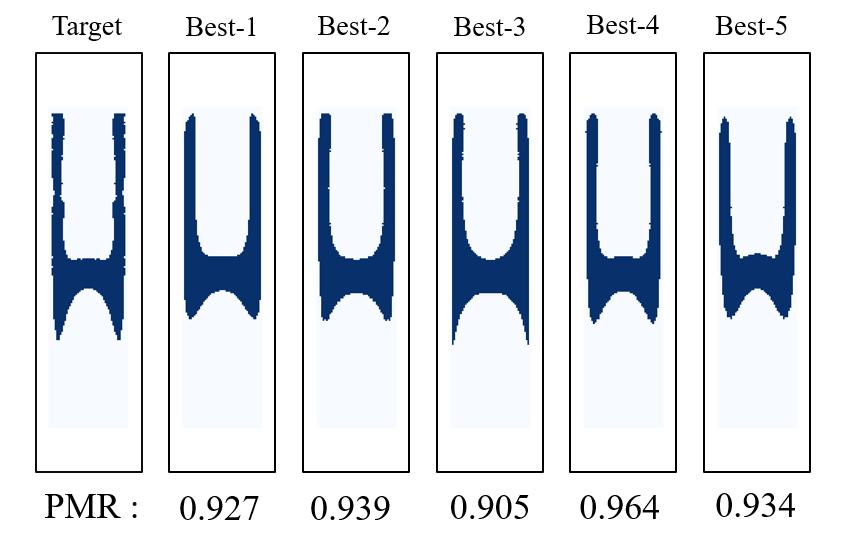}
 \caption{Top 5 solutions obtained by the RL agent for target flow shape 5. The corresponding pillar sequences of the top 5 solutions are shown in Table~\ref{Tab:Best 5}}
 \label{Fig:Targets}
\end{figure}

\begin{table}[h!]
\caption{Frequencies of best performing pillar sequences for a target flow shape 5: A target flow can be achieved with multiple design solutions. This table shows some of the possible design solutions which are highly frequently visited by the Agent.}
\begin{tabular}{l l l}
\hline
Method & Modality & Frequency \\
\hline
Target & 20, 14, 16, 23, 17, 4 , 6 & - \\

Best-1  & 30, 30, 24 & 50294  \\ 

Best-2  & 2, 29, 34, 31 & 20374 \\ 
 
Best-3 & 30, 31, 31 & 10431\\

Best-4  & 29, 31, 24 & 7861  \\

Best-5 & 30, 29, 25 & 6356  \\
\hline
\label{Tab:Best 5}
\end{tabular}
\end{table}

\begin{figure}
 \centering
 \includegraphics[width=0.5\linewidth,clip,trim={0in 0in 0in 0in}]{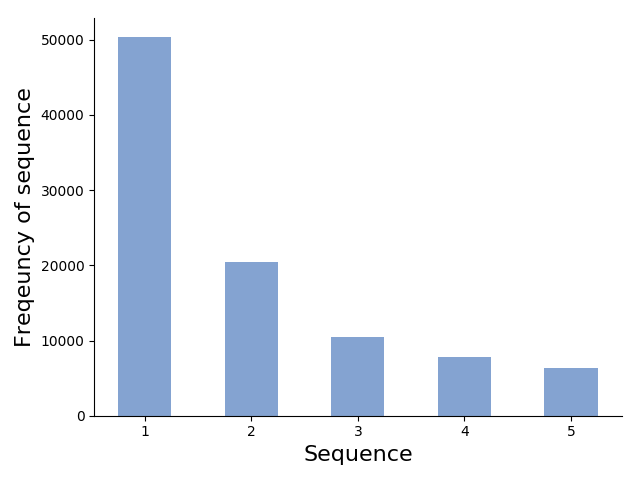}
 \caption{Bar chart illustrating the frequency in which the agent utilizes the top 5 solutions }
 \label{Fig:Targets}
\end{figure}
For the second formulation of the problem as a transfer learning problem, we first train the RL agent to reach target flow 7. After the agent has successfully learn to generate sequences for the flow, we use the trained agent and give it a new task of learning flow 6. We repeat the procedure above by using the agent that has been trained on 2 flows and train it again with a third and fourth flow. The results show that using an RL agent previously trained on a different flow enables the agent to learn the new flow shape much quicker than training the RL agent from scratch. It is also apparent that trained RL agent success frequency has significantly less variation as compared to the RL agent trained from scratch. This illustrates that it is possible to transfer the learning of an RL agent learning to solve one design problem to another design problem. Being able to transfer the knowledge of an RL agent has huge ramifications since it drastically reduces the amount of exploration or number of forward function evaluations required to arrive at a solution. Figure~\ref{Fig:TLComparison} shows the comparisons between a vanilla DDQN RL agent trying to learn 4 different flow shapes from scratch and an RL agent that has previously been trained on multiple flow shapes.  

\begin{figure}[h!]
 \centering
 \includegraphics[width=0.5\linewidth,clip,trim={0in 0in 0in 0in}]{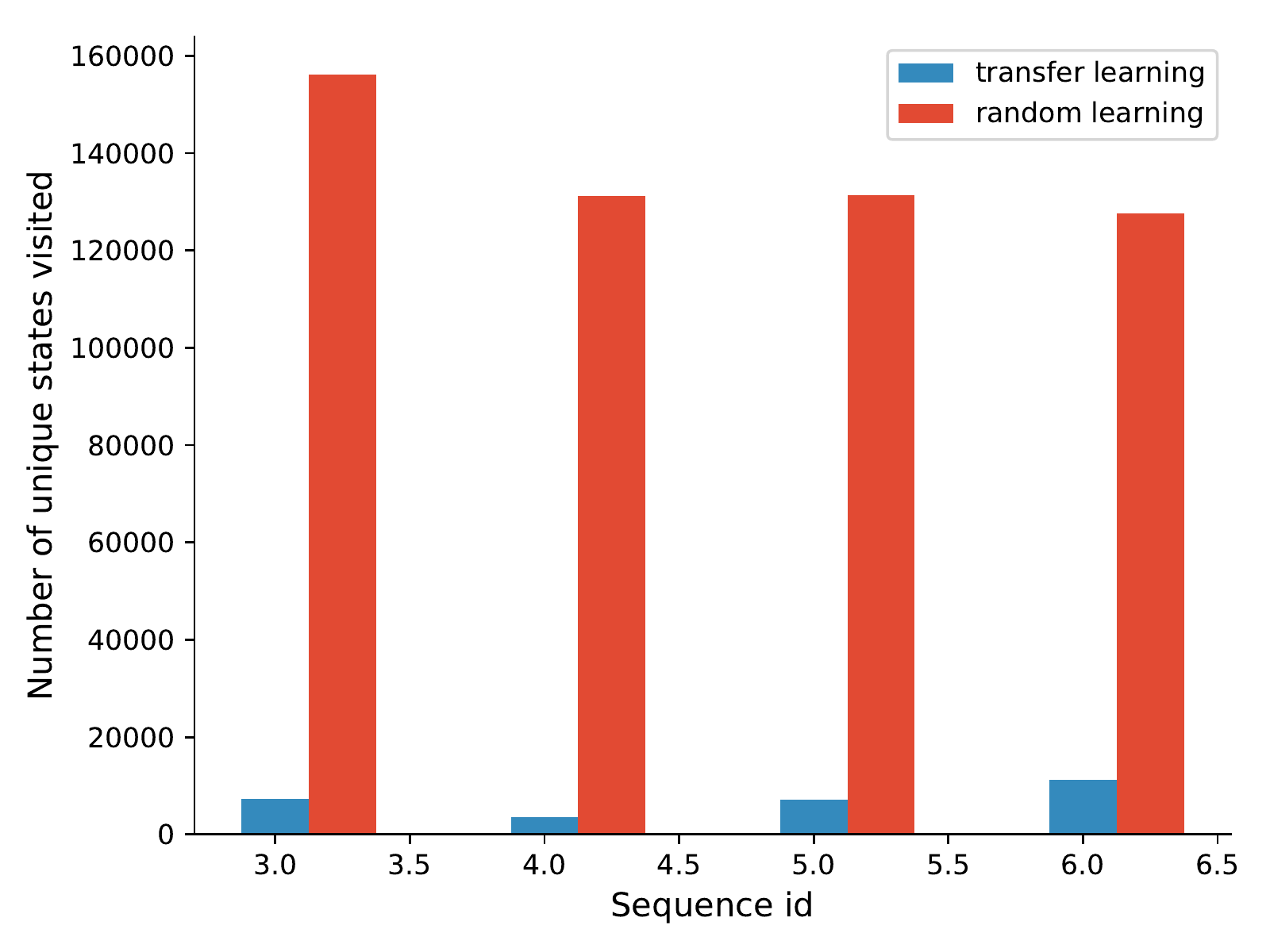}
 \caption{Bar chart illustrating the number of unique states visited by the agent while learning completely from random weights in compared to a agent using weights transferred from earlier learning.}
 \label{Fig:Targets}
\end{figure}

\begin{figure}[h!]
 \centering
 \includegraphics[width=1\linewidth,clip,trim={0in 0in 0in 0in}]{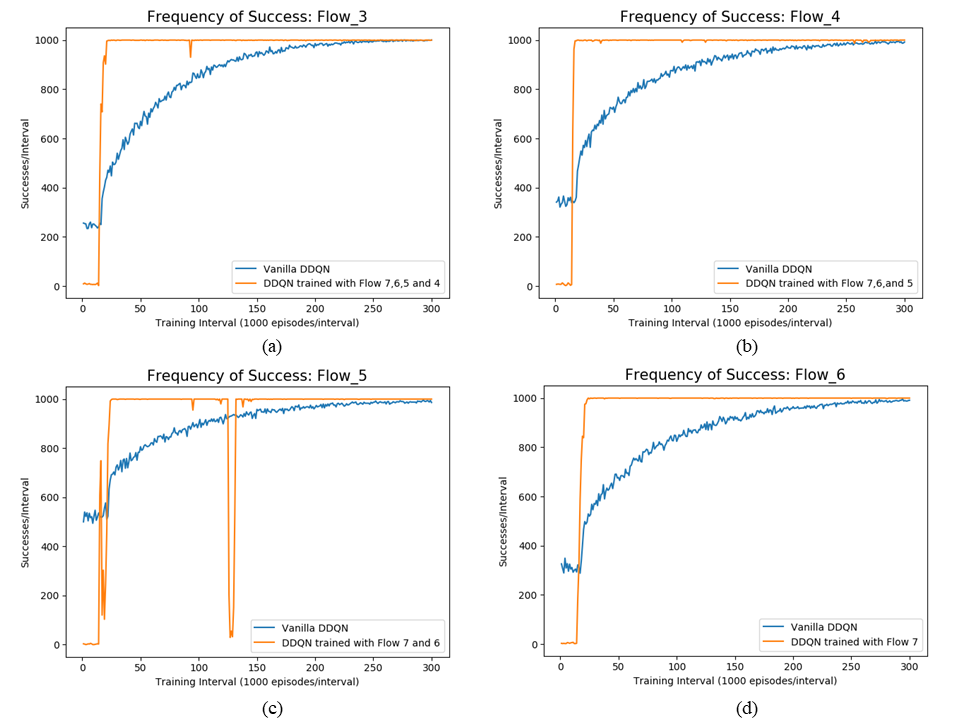}
 \caption{Performance comparisons between a vanilla DDQN RL agent learning a flow shape from scratch versus a RL agent that has previously been trained on other flow shapes}
 \label{Fig:TLComparison}
\end{figure}

\section{Conclusions}\label{sec:conclusions}
In this study, we propose a Deep Reinforcement Learning framework to solve an example of an inverse problem that could possibly have multiple solutions. We demonstrate that by formulating the inverse problem using RL, the RL agent is able to achieve a desired target. In particular, we build an RL agent for the inverse problem of designing pillar sequence for obtaining a target flow shape by learning multiple pillar sequences to arrive at a good approximation of the target flow shape. In addition, these sequences are also often shorter than the original target sequence, thus the solutions presented by the RL agent can be more optimal and cost efficient. In the second part of the study, we show that it is possible to speed up the learning of an RL agent by using an agent that has previously been trained on a different flow shape. This demonstrates the phenomena of transfer learning where the knowledge of one RL agent can be transferred to solve another set of inverse problem. Future efforts for this work include creating/using algorithms which are more sample efficient and are generalizable for any complexity of flow shape. \clearpage

\section{Acknowledgements}\label{sec:acknowledgments}
This work has been supported in part by the U.S. Air Force Office of Scientific Research under the YIP grant FA9550-17-1-0220. Any opinions, findings and conclusions or recommendations expressed in this publication are those of the authors and do not necessarily reflect the views of the sponsoring agency. We would also like to thank Sambit Ghadai for providing valuable insights and feedback on this work. 

\bibliographystyle{unsrt}
\bibliography{references}

\begin{thebibliography}{10}

\bibitem{Whitesides2006}
George~M Whitesides.
\newblock {The origins and the future of microfluidics.}
\newblock {\em Nature}, 442(7101):368--373, 2006.

\bibitem{Sackmann2014}
Eric~K Sackmann, Anna~L Fulton, and David~J Beebe.
\newblock {The present and future role of microfluidics in biomedical
  research.}
\newblock {\em Nature}, 507(7491):181--9, 2014.

\bibitem{Amini2014}
Hamed Amini, Wonhee Lee, and Dino {Di Carlo}.
\newblock {Inertial microfluidic physics}.
\newblock {\em Lab Chip}, 14(15):2739--2761, aug 2014.

\bibitem{amini2013engineering}
Hamed Amini, Elodie Sollier, Mahdokht Masaeli, Yu~Xie, Baskar
  Ganapathysubramanian, Howard~A Stone, and Dino Di~Carlo.
\newblock Engineering fluid flow using sequenced microstructures.
\newblock {\em Nature communications}, 4:1826, 2013.

\bibitem{labonachip}
Daniel Stoecklein, Chueh-Yu Wu, Keegan Owsley, Yu~Xie, Dino Di~Carlo, and
  Baskar Ganapathysubramanian.
\newblock Micropillar sequence designs for fundamental inertial flow
  transformations.
\newblock {\em Lab on a Chip}, 14(21):4197--4204, 2014.

\bibitem{Nunes2014}
Janine~K. Nunes, Chueh~Yu Wu, Hamed Amini, Keegan Owsley, Dino {Di Carlo}, and
  Howard~A Stone.
\newblock {Fabricating shaped microfibers with inertial microfluidics}.
\newblock {\em Advanced Materials}, 26:3712--3717, 2014.

\bibitem{Paulsen2015}
Kevin~S. Paulsen, Dino {Di Carlo}, and Aram~J. Chung.
\newblock {Optofluidic fabrication for 3D-shaped particles}.
\newblock {\em Nature Communications}, 6:6976, 2015.

\bibitem{Wu2015}
Chueh~Yu Wu, Keegan Owsley, and Dino {Di Carlo}.
\newblock {Rapid Software-Based Design and Optical Transient Liquid Molding of
  Microparticles}.
\newblock {\em Advanced Materials}, 27(48):7970--7978, 2015.

\bibitem{Paulsen2018}
Kevin~S. Paulsen, Yanxiang Deng, and Aram~J. Chung.
\newblock {DIY 3D Microparticle Generation from Next Generation Optofluidic
  Fabrication}.
\newblock {\em Advanced Science}, 5(7):1--6, 2018.

\bibitem{Sollier2015}
Elodie Sollier, Hamed Amini, Derek~E Go, Patrick~a. Sandoz, Keegan Owsley, and
  Dino {Di Carlo}.
\newblock {Inertial microfluidic programming of microparticle-laden flows for
  solution transfer around cells and particles}.
\newblock {\em Microfluidics and Nanofluidics}, pages 1--13, 2015.

\bibitem{Wu2018}
Chueh-Yu Wu, Daniel Stoecklein, Aditya Kommajosula, Jonathan Lin, Keegan
  Owsley, Baskar Ganapathysubramanian, and Dino {Di Carlo}.
\newblock {Shaped 3D microcarriers for adherent cell culture and analysis}.
\newblock {\em Microsystems {\&} Nanoengineering}, 4(1):21, 2018.

\bibitem{Du2008}
Yanan Du, Edward Lo, Shamsher Ali, and Ali Khademhosseini.
\newblock {Directed assembly of cell-laden microgels for fabrication of 3D
  tissue constructs.}
\newblock {\em Proceedings of the National Academy of Sciences of the United
  States of America}, 105(28):9522--7, 2008.

\bibitem{LeGoff2015}
Gaelle~C. {Le Goff}, Rathi~L. Srinivas, W.~Adam Hill, and Patrick~S. Doyle.
\newblock {Hydrogel microparticles for biosensing}.
\newblock {\em European Polymer Journal}, 72:386--412, 2015.

\bibitem{Yu2018}
Bing Yu, Hailin Cong, Qiaohong Peng, Chuantao Gu, Qi~Tang, Xiaodan Xu, Chao
  Tian, and Feng Zhai.
\newblock {Current status and future developments in preparation and
  application of nonspherical polymer particles}.
\newblock {\em Advances in Colloid and Interface Science}, 256:126--151, 2018.

\bibitem{Stoecklein2018}
Daniel Stoecklein, Keegan Owsley, Chueh-Yu Wu, Dino {Di Carlo}, and Baskar
  Ganapathysubramanian.
\newblock {uFlow: software for rational engineering of secondary flows in
  inertial microfluidic devices}.
\newblock {\em Microfluidics and Nanofluidics}, 22(7):74, jun 2018.

\bibitem{gaoptimization}
Daniel Stoecklein, Chueh-Yu Wu, Donghyuk Kim, Dino Di~Carlo, and Baskar
  Ganapathysubramanian.
\newblock Optimization of micropillar sequences for fluid flow sculpting.
\newblock {\em Physics of Fluids}, 28(1):012003, 2016.

\bibitem{DQN}
Volodymyr Mnih, Koray Kavukcuoglu, David Silver, Alex Graves, Ioannis
  Antonoglou, Daan Wierstra, and Martin~A. Riedmiller.
\newblock Playing atari with deep reinforcement learning.
\newblock {\em CoRR}, abs/1312.5602, 2013.

\bibitem{DDQN}
Hado van Hasselt, Arthur Guez, and David Silver.
\newblock Deep reinforcement learning with double q-learning.
\newblock {\em CoRR}, abs/1509.06461, 2015.

\bibitem{smccnn}
Kin~Gwn Lore, Daniel Stoecklein, Michael Davies, Baskar Ganapathysubramanian,
  and Soumik Sarkar.
\newblock Hierarchical feature extraction for efficient design of microfluidic
  flow patterns.
\newblock In {\em Feature Extraction: Modern Questions and Challenges}, pages
  213--225, 2015.

\bibitem{scirep}
Daniel Stoecklein, Kin~Gwn Lore, Michael Davies, Soumik Sarkar, and Baskar
  Ganapathysubramanian.
\newblock Deep learning for flow sculpting: Insights into efficient learning
  using scientific simulation data.
\newblock {\em Scientific reports}, 7:46368, 2017.

\bibitem{ppnitn}
Kin~Gwn Lore, Daniel Stoecklein, Michael Davies, Baskar Ganapathysubramanian,
  and Soumik Sarkar.
\newblock A deep learning framework for causal shape transformation.
\newblock {\em Neural Networks}, 98:305--317, 2018.

\bibitem{manochasiggraph18}
Pingchuan Ma, Yunsheng Tian, Zherong Pan, Bo~Ren, and Dinesh Manocha.
\newblock Fluid directed rigid body control using deep reinforcement learning.
\newblock {\em ACM Transactions on Graphics (TOG)}, 37(4):96, 2018.

\bibitem{stephanguy}
Bobby Davis, Nicholas Sohre, and Stephen~J Guy.
\newblock Multiworld motion planning.
\newblock {\em IEEE Robotics and Automation Letters}, 3(4):3968--3974, 2018.

\bibitem{rlvortexflow}
Jean Rabault, Ulysse Reglade, Nicolas Cerardi, Miroslav Kuchta, and Atle
  Jensen.
\newblock Deep reinforcement learning achieves flow control of the 2d karman
  vortex street.
\newblock {\em arXiv preprint arXiv:1808.10754}, 2018.

\bibitem{rlvortexflow2}
Rabault Jean, Kuchta Miroslav, and Jensen Atle.
\newblock Artificial neural networks trained through deep reinforcement
  learning discover control strategies for active flow control.
\newblock {\em arXiv preprint arXiv:1808.07664}, 2018.

\bibitem{rlgliding}
Guido Novati, Lakshminarayanan Mahadevan, and Petros Koumoutsakos.
\newblock Deep-reinforcement-learning for gliding and perching bodies.
\newblock {\em arXiv preprint arXiv:1807.03671}, 2018.

\bibitem{gym}
Greg Brockman, Vicki Cheung, Ludwig Pettersson, Jonas Schneider, John Schulman,
  Jie Tang, and Wojciech Zaremba.
\newblock Openai gym.
\newblock {\em CoRR}, abs/1606.01540, 2016.

\end{thebibliography}

\end{document}